\documentclass[twocolumn]{article}
\usepackage[utf8]{inputenc}

\usepackage{csquotes}
\usepackage{dirtytalk}
\usepackage{authblk}
\usepackage{graphicx}
\graphicspath{ {img/} }
\usepackage[dvipsnames]{xcolor}
\usepackage{amsmath}
\usepackage{amsfonts}

\title{In-flight Novelty Detection with Convolutional Neural Networks}
\author[1]{Adam Hartwell}
\author[1]{Felipe Montana}
\author[1]{Will Jacobs}
\author[1]{Visakan Kadirkamanathan}
\author[1]{Andrew R Mills}
\author[2]{Tom Clark}
\affil[1]{Department of Automatic Control and Systems Engineering, University of Sheffield, UK}
\affil[2]{Rolls-Royce Plc, UK}

\begin{document}
\maketitle

\section{Abstract}
    Gas turbine engines are complex machines that typically generate a vast amount of data, and require careful monitoring to allow for cost-effective preventative maintenance. In aerospace applications, returning all measured data to ground is prohibitively expensive, often causing useful, high value, data to be discarded. The ability to detect, prioritise, and return useful data in real-time is therefore vital. This paper proposes that system output measurements, described by a convolutional neural network model of normality, are prioritised in real-time for the attention of preventative maintenance decision makers.
    
    Due to the complexity of gas turbine engine time-varying behaviours, deriving accurate physical models is difficult, and often leads to models with low prediction accuracy and incompatibility with real-time execution. Data-driven modelling is a desirable alternative producing high accuracy, asset specific models without the need for derivation from first principles.
    
    We present a data-driven system for online detection and prioritisation of anomalous data. Biased data assessment deriving from novel operating conditions is avoided by uncertainty management integrated into the deep neural predictive model. Testing is performed on real and synthetic data, showing sensitivity to both real and synthetic faults. The system is capable of running in real-time on low-power embedded hardware and is currently in deployment on the Rolls-Royce Pearl 15 engine flight trials. 
    

\section{Introduction}
    Aerospace Gas turbine engines (GTE)s are complex machines that must be monitored and maintained in order to ensure reliable operation over long time periods \cite{liGasTurbinePerformance2009a}. A key concept to maximise reliability is condition based, rather than schedule based, maintenance \cite{zaidanGasTurbineEngine2016}. This requires an accurate assessment of engine condition, which can be challenging to obtain. 
    
    The condition of an engine is typically assessed relative to similar assets in a fleet, as well as functional performance thresholds, and requires expert engineering knowledge of the maintenance decision makers. The finite human resource to assess engine performance is traditionally supported by small data snapshots and failure-mode specific feature engineering.  Larger data packets are collected from in-service on an ad-hoc basis, but this is logistically costly and often limited to cases where operational disruption has been observed. New methodologies are needed to support more efficient in-service operations.
    
    Data, obtained from a large variety of different sensors placed on the engine, is the main resource available to assess engine condition. However, the large number of sensors, in-flight bandwidth limitations, and on-board storage limitations mean that it is not possible to return all data to ground for analysis in most current systems \cite{hofmannMassiveDataTransfer2021}. It is, therefore, necessary to prioritise which data is returned and, in the context of condition based maintenance, prioritise data that is anomalous as this may indicate issues that require maintenance \cite{zhaoReviewGasTurbine2016a}. Furthermore, computational and data storage resources are extremely limited on the current generation of civil aerospace GTEs due to cost and weight restrictions and the harsh operating environment. Any data prioritisation system must function in real-time with limited computational resources lest data become unavailable before it can be saved. 
46

    Modern detection schemes make use of real-time engine models that run on-board in parallel to the real system, recently referred to as a Digital Twin \cite{glaessgenDigitalTwinParadigm2012a, taoDigitalTwinIndustry2018a}. Model-based schemes have the advantage of not requiring fault data to validate their performance, and assessments of nominal model accuracy determine the detectability of an asset's condition divergence from its twin. Such models are traditionally derived from physical principals by domain experts. Physical models have advantages, particularly in terms of explicability and their ability to extrapolate (when designed correctly). However, they are are costly to develop, difficult to tune to an individual asset, typically bespoke, and are computationally expensive to run \cite{asgariModellingSimulationGas2013a, panovGasturbolibSimulinkLibrary2009a}. An alternative paradigm is data-driven modelling, where a flexible model structure is fitted to the system by training on historical data. Data-driven models are often poor in terms of explicability and their ability to extrapolate beyond the training set - but can be more accurate in prediction for individual assets with a smaller computational demands relative to a physical model.     
    
    There is, therefore, a need for an anomaly detection and prioritisation system that is capable of running on-board under strict computational resource limitations. To address this need  we present a data-driven system that is capable of real-time anomaly detection and prioritisation on embedded hardware. The system can be used on subsets of the GTE system, or on the system as a whole, facilitating the prioritisation of high value data for condition monitoring. This system is currently in deployment on Pearl 15 engines.
    
    The developed system is based on the Digital Twin paradigm, that is, accurate simulation of an individual 'as-operated' gas turbine. Simulation of the whole gas turbine is infeasible given the computational constraints of embedded devices, however by selecting interesting input-output subsets it is possible to generate many data-driven models to form a Digital Twin. Each model may then be run in parallel to give wide coverage of the gas turbine, allowing for generalised anomaly detection without excessive computational overhead.
    
    Each model is trained using nominal data to make a prediction about a single signal given its input subset. The model also predicts a confidence on its prediction based on the model input. Model inputs that are not well described by the training data lead to a low confidence - thus allowing the distinction between model errors that arise from the unusual data of interest, and those from previously unseen operating conditions \cite{nixEstimatingMeanVariance1994a}. The system assumes that a sufficient amount of nominal data that covers expected operating conditions is available at training time. 
    
    The difference between model prediction and measurement, termed the residual, has been widely researched as a fault detection technique. The difficulty in obtaining an accurate model for the plant led to significant early work \cite{willskyiSurveyDesignMethods} on the development of diagnostic methods with plant observers (using data measurements to adapt state estimates), and with parity equation approaches decoupling known disturbances from faults into model subspaces \cite{frankFrequencyDomainApproach1994, friskRobustResidualGeneration2006}. Such methods rely on analytical forms of the plant, faults and expected disturbances. The difficulties in suitably modeling gas turbine behaviour (e.g. its non-linear nature and inter-unit variance) have led to these approaches being augmented with data-driven ‘tuning’ elements \cite{brothertonESTORMEnhancedSelf2003a, kobayashiHybridKalmanFilter2008} at the cost of increased complexity. Machine learning offers a more generic, lower cost, and faster to market approach, as well as the potential to accurately capture modelled physics behaviours embedded in the data.

    Other researchers have explored full or partial gas turbine anomaly detection using the full gamut of machine learning techniques \cite{zhaoReviewGasTurbine2016a}. Deep learning techniques have also been of particular relevance to the area in recent years \cite{amozegarEnsembleDynamicNeural2016a, zhongNovelAnomalyDetection2020a, yanAccurateReliableAnomaly2019b, yanDetectingGasTurbine2020b, leeUnsupervisedAnomalyDetection2020a, liuFaultDetectionGas2018a, luoGasTurbineEngine2017a} due to their success in other fields. These solutions are not suited to deployment on resource-limited hardware due to their high computational cost at run-time and/or their memory requirements.

    Anomaly detection using limited hardware has also been explored, using remote processing \cite{amontamavutSeparatedLinuxProcess2012a}, look-up tables \cite{summervilleUltralightweightDeepPacket2015a}, model-based comparison \cite{attiaOndeviceAnomalyDetection2015a}, and in other heavily band-limited system \cite{furanoUseArtificialIntelligence2020}. These solutions, however, are not suited to gas turbine anomaly detection due to the complexity of modelling the highly non-linear asset behaviours.
    
    Our system's novelty is derived from its combination of a flexible modelling approach (allowing prediction of any input-output path within a gas turbine), a confidence prediction for quantifying uncertainty in the models, and software that allows for real-time operation on embedded hardware. Furthermore, our solution is especially flexible as it allows drop-in, and drop-out of models without retraining.

\section{Methods}
    We use Chandola \textit{et al's} \cite{chandolaAnomalyDetectionSurvey2009a} definition of an anomaly:
        
    \say{Anomalies are patterns in data that do not conform to a well defined notion of normal behavior}
    
    That is, we do not know how anomalies will present themselves ahead of time and therefore must instead define normal in order to determine when anomalies occur. We therefore take the data-driven Digital Twin approach of selecting a number of inputs and an output based on engineering knowledge, so that the error between the model prediction and true signal value may be evaluated to determine whether data is normal.
    
    This viewpoint is also particularly useful in the gas turbine context because their high reliability means that the vast majority of available data is of nominal operation. A prediction approach allows a proxy measurement for detection in the form of prediction error. This is useful because it provides an indicator of model performance even when there is limited data with faults available.
    
    Each model is designed to predict only a single output, multiple models with different input-output subsets are then required to run in parallel to provide wide coverage of GTE operation. This allows each model to be made lean and computationally efficient while simultaneously providing a simple way to scale to the computational resources available.
    
    The complexity of gas turbines coupled with the variety of operating conditions and environmental factors that affect their operation presents an additional problem for anomaly detection: data acquisition. It is difficult to acquire enough data to well define normal for all possible conditions, environments, and modes of operation. Therefore, we designed our models following the advice of Nix and Weigend \cite{nixEstimatingMeanVariance1994a}, so that each model predicts a probability distribution characterised by a mean and confidence, see Section \ref{sec:nn_arch}.
        
    \subsection{Anomaly Detection System}
        The architecture of the Anomaly Detection System is shown in Figure \ref{fig:detector_system} which allows for scalable, real-time operation:
        
        \begin{figure}[ht]
            \centering
            \includegraphics[width=0.43\textwidth]{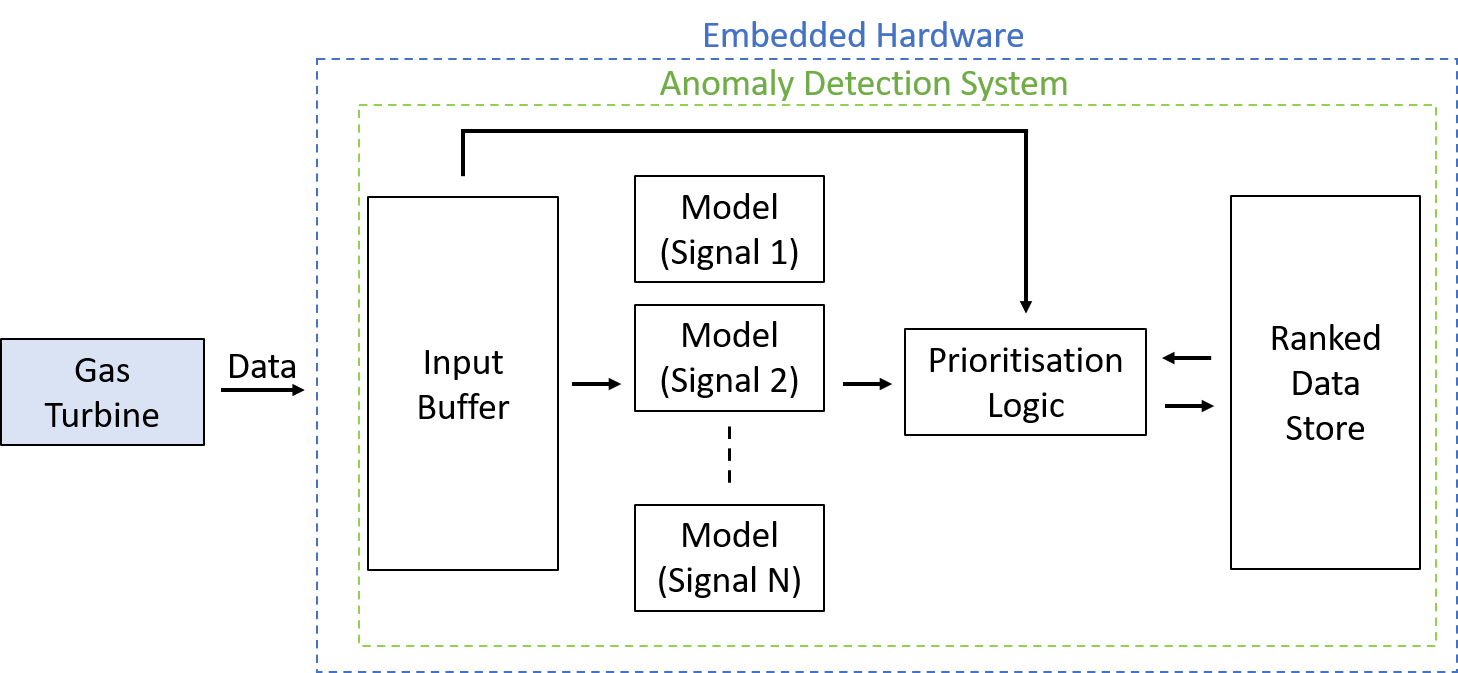}
            \caption{System architecture. Data from the gas turbine is piped in and buffered before being run through the currently loaded models. The most unusual data is then stored ready for later review.}
            \label{fig:detector_system}
        \end{figure}
        
        The architecture allows streaming of data from sensors attached to a gas turbine in real-time into a local input buffer. The buffer is then used to feed $N$ models, where $N$ is selected based on a trade-off of coverage with computational resources. Data prioritisation is performed based on the Standardised Euclidean Distance metric described in Equation \ref{eq:distance} (Section \ref{sec:data_priority}) and the most anomalous data stored in a local data store. The data stored can be any sensor information from the gas turbine and need not be limited to a model input-output subset.
        
        Data in the store may be accessed whenever a data link is available, for example, via satellite link when in-flight, or airport connectivity when landed. The store may be sized to match memory requirements of the hardware.
    
    \subsection{Data modelling} \label{sec:data_modelling}
        The anomaly detection scheme presented in the previous section is reliant on the existence of models that can map observations of the input data to observations of output data.  Consider a single input-output data set  $\mathcal{D}=\{X, \mathbf{y}\}$, where
        \begin{align*}
            \mathbf{y} &= [y_1, y_2, \ldots, y_N]^T \in \mathbb{R}^{N,1}, \\
            X &= 
        \end{align*}   
        where $N$ is the length of the data series and $m$ is the number of channels in the input data. The input-output mapping is approximated with a function $f(\cdot)$, such that the $n$'th element of the output can be modeled as
        \begin{equation}
            y_n = f(\mathbf{x}_n) + e(\mathbf{x}_n), \label{eq:function}
        \end{equation}
        where $e(\cdot)$ is additive noise that is dependant on the input data. 

        The estimation of the distribution of the output data is central to the anomaly detection scheme, allowing the quantification of uncertainty on the estimate. Importantly, the uncertainty is a function of the input data $\mathbf{x}_n$ to quantify the uncertainty in the model in different regions of the input space. The output distribution is described by a multivariate Gaussian with mean $\mu_n=\mu(\mathbf{x}_n)$ and variance $\sigma_n = \sigma(\mathbf{x}_n)$ given by
        \begin{equation}
            \mathcal{N} \left( y_n | \mu_n, \sigma^2_n\right) = \frac{1}{{ \sqrt {2\pi \sigma^2_n} }}e^ {-\frac{(y_n - \mu_n)^2 } {2\sigma^2_n }} . \label{eq:Normal_distribution}
        \end{equation}
        
        In this work, the input dependant mean and variance are estimated simultaneously using a deep neural network, the design, training, and real-time implementation of which are discussed in the remainder of this section.

    \subsubsection{Neural Network Design} \label{sec:nn_arch}
        The choice of a recurrent neural is often favoured when dealing with time-series data, however, their high computational and memory costs make them generally unsuitable to implement on current embedded hardware \cite{rezkRecurrentNeuralNetworks2020a}. We therefore chose to use a convolutional Neural Network architecture in order to minimise the computational cost of the forward passes.
        
        The large number of sensors and complexity of the gas turbine lead to our decision to use a flexible network design that can be adapted to any target input-output data set. Further we also engineered the design to be flexible in terms of the number of time-steps the network can take as an input. Typically this requires a large number of extra weights to scale the number of time-steps. Our design uses a constrained convolutional layer (the Temporal Fold) to minimise the computational impact of inputting extra time-steps. This is especially important as, while some input-output subsets are expected to have short time-lag relationships, others e.g. temperatures, may have much longer time constants which a convolutional network will not be able to represent if the input data is unavailable.
        
\begin{figure}[ht] 
            \centering
            \includegraphics[width=0.4\textwidth]{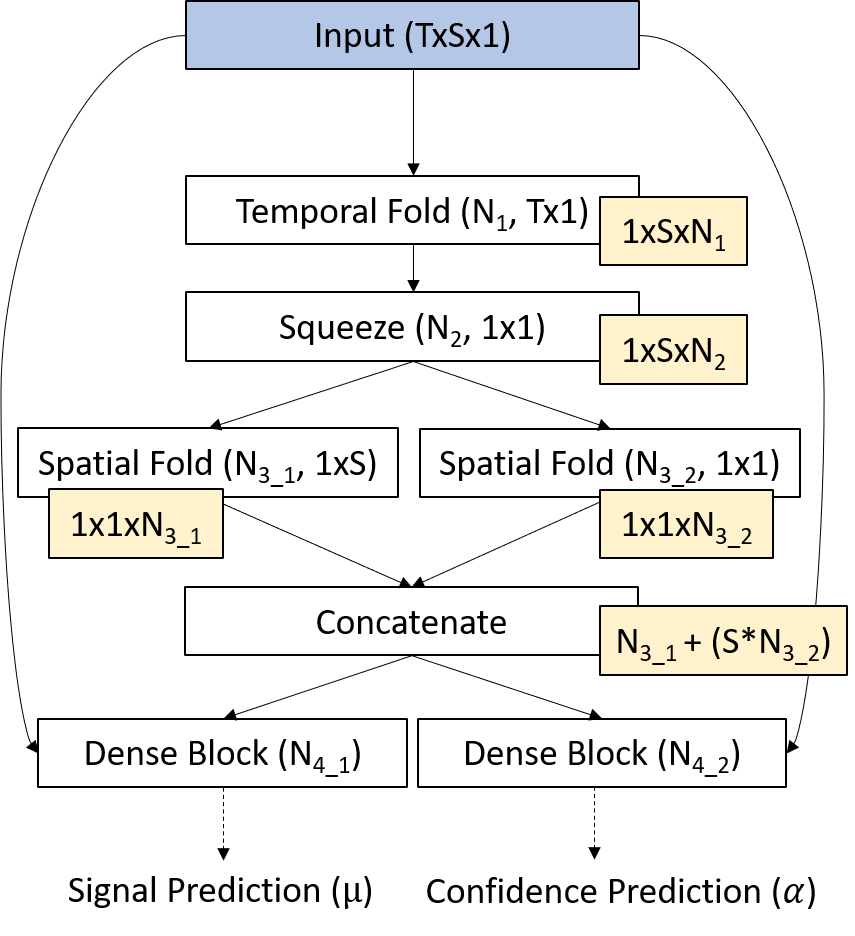}
            \caption{Overview of the network architecture. Number of hidden units and stride size given in brackets, output size given in yellow. The design is flexible such that \enquote{T} matches the number of incoming timesteps in each window and \enquote{S} is the number of signals. The \enquote{Dense Blocks} are 3 dense layers each followed by a dropout layer with $rate=0.5$. The number of hidden units ($N_{X}$) in different layers may be adjusted based on the complexity of the input-output set and available computation.}
            \label{fig:nn_arch_abstract}
        \end{figure}

        The design is flexible in terms of the number of inputs, which allows for deployment on new data by non-specialists; since no redesign of the architecture is necessary to accommodate for different input-output subsets or timescales. The network architecture takes advantage of skip connections to improve performance at minimal computational cost, dropout regularisation to reduce over-fitting \cite{srivastavaDropoutSimpleWay2014a}, and Leaky Rectified Linear Units (LReLU) throughout. The conceptual architecture is shown in Figure \ref{fig:nn_arch_abstract} which is described in more detail below;
        
        \textbf{Input} Incoming data is treated as a 2D window with length $T$ and width $S$, where $T$ and $S$ are time steps and the number of signals for a particular input-output subset respectively. This simplifies implementation and visualisation as it means the data can be treated in a similar manner to images.
        
        \textbf{Temporal Fold}
        The stage of the network is a 1D convolution in the time dimension, explicitly enforcing feature extraction from individual signals at the lowest level, here named the Temporal Fold. Denoting the convolution operator as $*$, the discrete 1D convolution operation is defined as;

        \begin{equation} \label{Eq:conv 1D}
            s_j^{[l]} = (\mathbf{z}^{[l-1]}_j*\mathbf{K}) = \sum_{i} \mathbf{z}^{[l-1]}_j(i)\mathbf{K}(i)
        \end{equation}  
        where $\mathbf{z}^{[l-1]}_j$ is the output of the $j$'th neuron at the $l$'th layer of the network and $\mathbf{K}$ is a 1D kernel of appropriate size. The temporal fold takes the network input as input $\mathbf{z}^{[l-1]}_j(i) = \mathbf{x}_{k-T:k}$ with $\mathbf{K}=\mathbf{K}_{TF}$ is the Temporal Folder Kernel with with height $T$, width $1$, and depth equal to the number of input features. The output of the convolution can be seen as a weighted average over the past $T$ data points for each input channel, where the weighting is defined by the elements of $\mathbf{K}_{TF}$. 
        
        The output of the convolution is passed though a LReLU activation function defined as,
        \begin{align}
            z^{[l]}_j &= g(s_j^{[l]}) \\
            g(s_j^{[l]}) &=
            \begin{cases}
                s_j^{[l]} & \text{if } s_j^{[l]}>0, \\
                0.01s_j^{[l]} & \text{otherwise}
            \end{cases}  \label{Eq:ReLu}
        \end{align}
        
        \textbf{Squeeze}
        The next stage of the network is \enquote{Squeeze} layer. It consists of a $1\times 1$ Squeeze convolution which outputs to a 1xS convolution and a 1x1 convolution, both described by Equation \ref{Eq:conv 1D} with appropriately sized kernels and with LReLU activation functions described by Equation \ref{Eq:ReLu}. This method is inspired by Squeezenet \cite{iandolaSqueezeNetAlexNetlevelAccuracy2016a}, and replaces the $c\times c$ convolution for a $1\times S$ convolution in order to significantly reduce the number of hyper-parameters to be learned.         
        
        \textbf{Spatial Fold}
        The two outputs of the Squeeze layer are passed through a 1D convolution described by Equation \ref{Eq:conv 1D}, this time in the spatial dimension (across features generated at the previous layer) and thus named the Spatial Fold. A LReLU activation function is again used. The layer determines cross-channel features and inter-filter features. The outputs of the convolutions are flattened and concatenated. 
        
        \textbf{Dense layers - Mean Prediction:} A skip connection is performed, where the output of the Spatial Fold is concatenated with the flattened original input. The concatenated outputs are then passed into a series of three dense layers described by;
        \begin{equation}
            z^{[l]}_j = g(W z^{[l-1]}_j + \mathbf{b})
        \end{equation}
        
        where $W$ and $\mathbf{b}$ are hyper-parameters to be learned and $g$ is the $L-ReLU$ activation function defined by Equation \ref{Eq:ReLu}. A dropout layer with $rate=0.5$ is placed after each dense layer to reduce over-fitting during training. The output is a single value that is an  mean estimate of the output for the given network input.
        
        \textbf{Dense layers - Confidence Prediction:} A further skip connection is performed where, again, the output of the Spatial Fold is concatenated with the flattened original input but here the mean estimate is also concatenated. This is then passed through into a series of three dense layers, as above for the mean estimation, and outputs an estimate of the output variance. 
        
    \subsubsection{Neural Network Optimisation} \label{sec:nn_opt}    
        Optimisation of the model weights of a deep neural network requires an optimisation target. Considering the model described by Equation \ref{eq:Normal_distribution}, it is simple to construct a likelihood function as 
        \begin{equation}
            \mathcal{LL} = \prod_{n=1}^B \mathcal{N}\left( y_n |\mu_n, \sigma^2_n\right),
        \end{equation}
        where $B$ is the batch size. Maximisation of the likelihood is equivalent to the minimisation of the negative log-likelihood 
        \begin{align} 
            \mathcal{NLL} &= -\frac{1}{B} \sum_{n=1}^{B} \ln \Big( \frac{1}{\sqrt{2 \pi \sigma_n^2}} e^{- \frac{1}{2} \Big( \frac{(y_n-\mu_n)^2}{\sigma_n^2} \Big)} \Big) \\
            &=\frac{1}{B} \sum_{n=1}^{B}  \Big( \frac{1}{2} \Big( \frac{(y_n - \mu_n)^{2}}{\sigma_n^2} \Big) -  \ln(\sigma_n) - \frac{1}{2} ln(2 \pi) \Big)  \label{eq:loss}
        \end{align}
        which is used as the target for optimisation \cite{nixEstimatingMeanVariance1994a}.
        
        Optimisation was performed using the Rectified ADAM (RADAM) algorithm \cite{kingmaAdamMethodStochastic2014a, liuVarianceAdaptiveLearning2019a} combined with early stopping on a separate validation data split. The parameters of the Rectified ADAM algorithm were varied depending on the input-output subset based on experiments with the validation set. RADAM was selected based on it's performance relative to a small pool of candidate optimisers.
        
        The network outputs the mean, $\mu_n$, and the confidence $\alpha_n = \ln(\sigma_n)$. The re-parameterisation enforces $\sigma_n>0$ to ensure that the network output is always valid. $\sigma_n$ is regenerated after the forward pass using $\sigma_n = e^{\alpha_n}$. The assumption of normally distributed residuals is reasonable, given that we expect random fluctuations from sensor readings. 
        
        This confidence estimation approach is based on the work of Nix and Weigend \cite{nixEstimatingMeanVariance1994a} and similar to more recent work e.g. bounding box estimation in computer vision \cite{redmonYouOnlyLook2016}). Other approaches to uncertainty estimation in neural networks include using dropout at run-time \cite{galDropoutBayesianApproximation2016}, using ensembles as a prediction scatter and Bayesian neural network solutions. We utilise the explicit estimation approach since it allows us to enforce a Gaussian confidence estimation while retaining the greater flexibility of non-Bayesian neural networks and fast run-time of a single network solution.
        
        
    \subsubsection{Model and Hyper-parameter Selection} 
        In order to ensure a neural network was an appropriate model choice, a number of simpler models were tested as a benchmark. These models include Lasso, Ridge, Elastic Net linear regression and Regression Forests. Each was found to perform worse than the neural network solution and naturally lacked the ability to estimate confidence. Typically it was possible to train these models to work well at a local scale, performing the same or better than the neural network over small portions of the data, but considerably worse elsewhere indicating a tendency to over-fit to the training data and a lack of generalisation.
        
        Network hyper-parameters were selected based on starting values from the literature followed by jittered grid search and manual tuning. Validation set (not test set) data  was used to evaluate performance and inform selection. More information on the hyper-parameters are included in Appendix \ref{sec:app1}.
        
    \subsubsection{Model Comparison and Data Prioritisation} \label{sec:data_priority}

        The system-level assessment of abnormality relies on being able to compare measurements to the outputs of any number of different models where each model may have a different output range. This is achieved by treating outputs of the network as a multivariate Gaussian and adopting the Mahalanobis distance \cite{demaesschalckMahalanobisDistance2000a}. Since each model is independent, such that the co-variance matrix is diagonal, the Mahalanobis distance reduces to Standardised Euclidean Distance (SED). We then refactor the equation for novelty ranking and drop the square root since it is unnecessary for calculating the relative ranking of each sample. This leaves Equation (\ref{eq:distance}) that retains the intended functionality, but can be computed quickly in an online context,
        
        \begin{equation} \label{eq:distance}
            \tilde{d}(y_n,\mu_n)= \frac{{(y_n-\mu_n)^{2}}}{\sigma_n^{2}}, \sigma = e^{\alpha}
        \end{equation}
        
        This setup relies on models being up-to-date, that is, if an engine maintenance is performed then it is reasonable to expect some input-output relationships to change. Therefore, an out-of-date model may correctly indicate anomalous data for its input-output set, however, this is likely to be undesirable in practice. Methods for keeping models up-to-date are beyond the scope of this paper, and are discussed in our forthcoming work.

    
    \subsubsection{Real-time Implementation}
        Deployment to embedded hardware was achieved via an application written in C++. The Tensorflow Lite \cite{abadiTensorflowSystemLargescale2016a} library was employed to ensure fast forward-passes of all models involved while the application took care of the input and output buffering as well as data prioritisation tasks.
        
        Design assurance for the application was achieved via a combination of automated code analysis (via CLang Format), automated testing, memory profiling (via Valgrind) and static analysis. 
        
        In order to evaluate performance, internal timers were used around the critical code sections in order to review data load and data evaluation times (model runs combined with data prioritisation). This allowed monitoring of real-world performance.
    
    \section{Data Sets} 
        Two different data sets were used to validate that the system produced appropriate predictions and gave acceptable novelty detection performance.
        
        Each data set was split into three sets: Training, Validation, and Test. Test data was used solely for evaluation and, unless otherwise specified, consisted of the most recent data. Training and Validation data was drawn from earlier data randomly. This split setup mimics real world applications where only past data is available and models must be deployed to an aircraft. In all cases, the Training set was used for determination of scaling and the only data directly available to the networks for learning. The Validation set was used for early stopping and for hyper-parameter tuning while the Test set was kept solely for evaluation.
        
        \textbf{NASA TurboFan Degradation Dataset} \cite{saxenaDamagePropagationModeling2008a}. This data comes from a prognostics competition aimed at assessing the quality of remaining useful life prediction algorithms. Specifically, subset \enquote{FD002} was used, that contains data on synthetically modelled degradation over six different operating conditions (combinations of altitude, air speed and thrust settings). The data simulates one measurement snapshot of each measured signal per flight to characterize an engine's performance during that flight. For the selected data set, 519 (split between training and testing sets) different engines with different initial health conditions were simulated over increasing synthetic degradation, each engine is termed a simulation 'run'. Our training data (marked for testing in the original) contained only low degradation and no failures, while the test set (marked for training in the original) had progressed to failure conditions near the end.  Further, only the first half of each run was used during model training to help prevent the model learning the degradation.
        
        \textbf{In-Service Fault Case Study}. This case study, provided by Rolls-Royce, includes continuous data from a twin engine aircraft across 116 flights at 1Hz.  This data, in contrast to the NASA data set, contains engine dynamics and real world disturbances, but a relatively low level of degradation and only one engine. There is one known anomaly in this data (validated by domain experts) as well as several other potential anomalies. Only one engine from the aircraft is used for modelling so to represent the real aircraft avionics, which do not allow real-time access to signals from both engines by design. The data presented here has been normalised and anonymised to protect the interests of parties that supplied it.
        
        

\section{Results and Discussion} \label{sec:results}
    \subsection{NASA TurboFan Degradation}
        For each sensor signal in the data a model was trained using the three \enquote{operating condition} signals as input in order to ensure a model of nominal operation was learnt. Since the data consists of one measurement snapshot per flight, the window length was set to 1 i.e. only considering most recent data and without information on engine physical dynamics.

\begin{figure*}[tbp] 
            \centering
        \includegraphics[width=\textwidth]{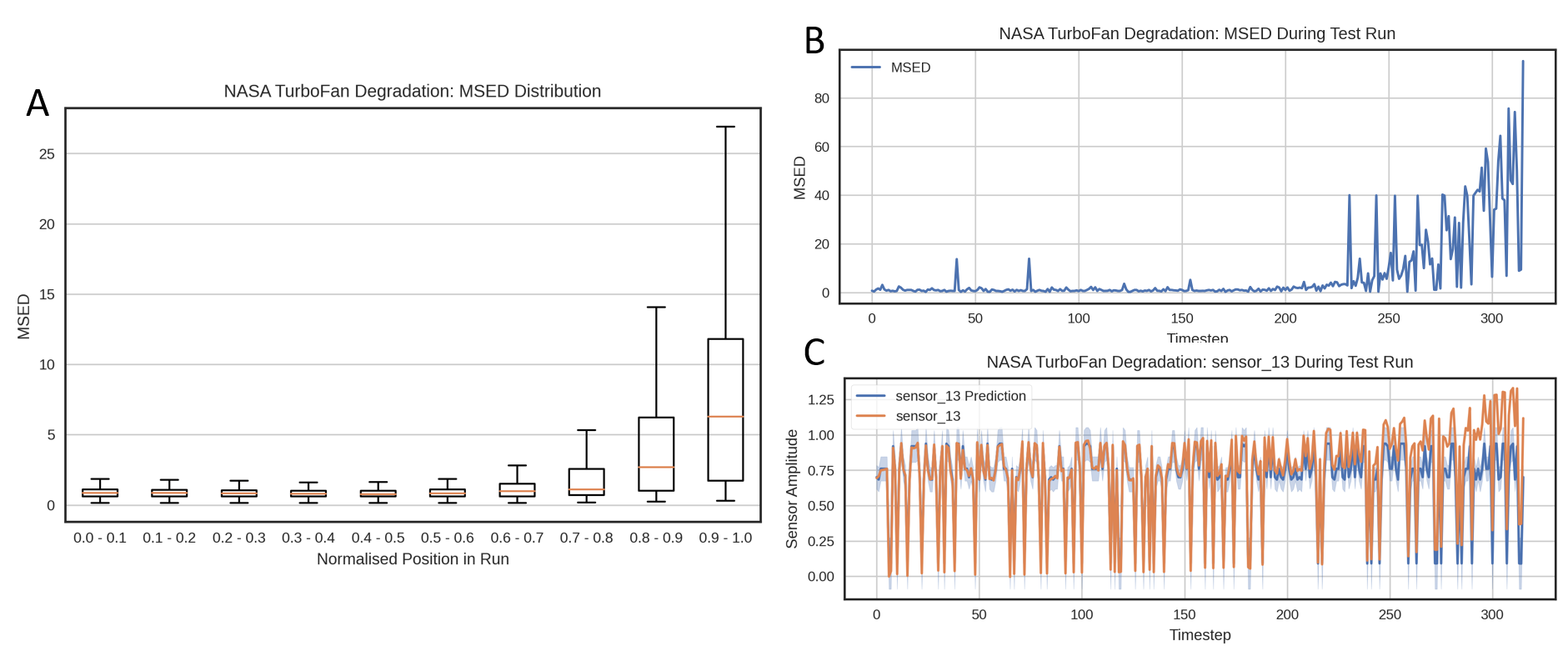}
            \caption{A: Distribution of MSEDs based on normalised position within test run demonstrating higher mean and spread towards the end of runs where faults occur. B: Example distribution of MSED over a test run demonstrating much higher values in the latter section where a fault occurs. C: Prediction for dominant (highest SED) model during the run, changes in shaded 95\% confidence interval demonstrate cause of MSED oscillation is due to lack of confidence.}
            \label{fig:nasa_box}
            \label{fig:nasa_dists}
        \end{figure*}

        The model was validated with a Mean Absolute Error (MAE) of 0.029 +/- 0.034 (normalised range) across all sensors on all the engines in the validation data. The growth in plant-model mismatch, measured by Mean Standardised Euclidean Distance (MSED), is due to the increase in degradation towards the end of each run, as observed in Figure \ref{fig:nasa_dists} A.
        
        
        The high variance in the MSED signal (see Figure \ref{fig:nasa_dists} A) is due to changes in the estimated confidence, which is similar across all signal models due to them using the same three operating condition inputs. The operating conditions in the MSED troughs of Figure \ref{fig:nasa_dists} B are different to those found in the training data which generates a wide confidence interval (since the model is less likely to be a good predictor) and therefore returns a small NSED even if the prediction error is relatively high. Therefore, the model correctly indicates that its predictions may be less accurate in these regions. 
        
        In the test data, it is known that the Turbofan degrades to a failure state at the end of each run.  It is therefore expected (and desirable) that the model will better represent the data at the start of the run and diverge from the plant behaviour at the end of each run. In order to test this property, the distribution of the MSED is compared for the 1st and 2nd half of each run contained in the test data, for each turbofan. 
        
        
        If the system is functioning correctly, the average MSED over the second half of each run would be greater indicating successful prioritisation of degraded (more unusual) data. The distributions were compared using a Mann-Whitney U test (since there is no guarantee of normality) under the null hypothesis that both sets came from the same distribution. The arithmetic means of the distributions were then compared, a larger deviation from 0 indicating more unusual data.
        
        The two sets of distributions where determined to have different shapes with a p value $\approx 0$ ($<1e^{-308}$). The mean of the earlier portion of the runs was $0.980$ while the latter portion was $3.466$ thus demonstrating that the system correctly prioritises this data. 
        

        Not all runs are as clear-cut, however, as demonstrated by the Mann-Whitney U test, the trend of having higher distances near to faults is consistent. 

        
        Each model in this setup used 74,946 parameters making it suitable for deployment in real-time on embedded hardware (see Section \ref{sec:realtime}).
        
        These results illustrate that the modelling process and prioritisation methodology are applicable to a time series of static snapshot generated by a time-varying process\footnote{The code to reproduce these models and results on the NASA data-set is made available alongside this paper.}. The model flexibility allows the approach to be applied without structural changes to dynamic data, as shown in the next section. 
        
    \subsection{In-Service Fault Case Study} \label{sec:res_fault_case}
\begin{figure}[hbtp] 
            \includegraphics[width=0.49\textwidth]{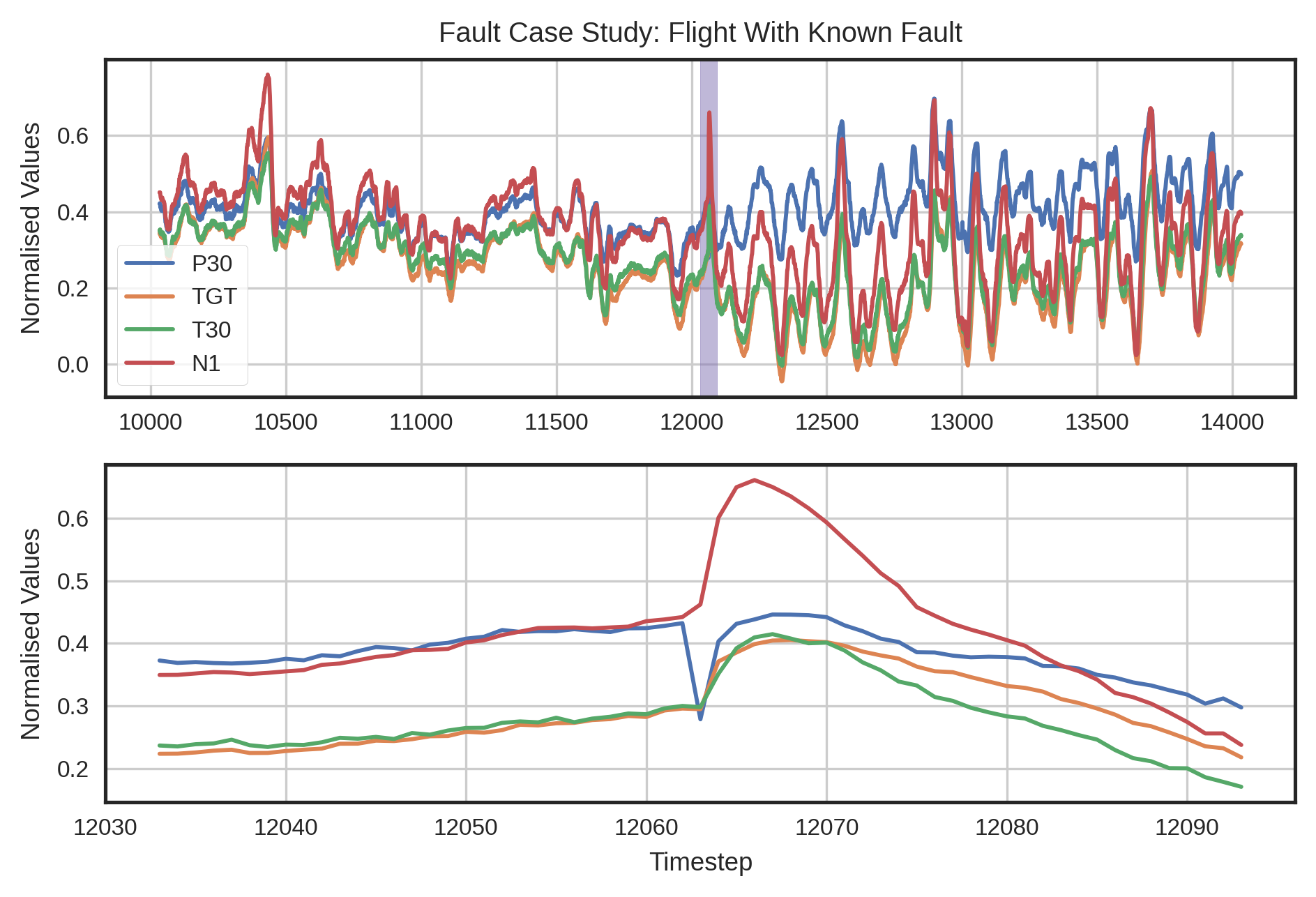}
            \caption{Normalised signals for the flight with the confirmed fault. The highlighted region in the top graph is zoomed in on in bottom graph. The spike in the internal pressure signal (P30) in the bottom graph is a known fault as is all behaviour afterwards. The temperature T30 is an internal temperature detected in the same locale as P30, N1 is the low speed shaft, and TGT is the Turbine Gas Temperature.}
            \label{fig:p30_anomaly}
        \end{figure}

         This case study concerns the detection of a real fault that was detected in service and has been confirmed by other health monitoring systems and domain experts at our industrial partner. The fault was detectable in real-time using the current generation of health monitoring techniques, but no warning was given. The symptoms were noted as being most prominent in the P30 reading and related systems. The initial spike is most prominent, although not unique in the dataset, and after the spike, engine behaviour changes in an anomalous way for the rest of the flight, see Figure \ref{fig:p30_anomaly}.
         
        
        A model was trained using the scheme described in Section \ref{sec:nn_arch} for each of the four signals shown in Figure \ref{fig:p30_anomaly}. The input to each model are the signals not predicted for, see Appendix \ref{sec:app2} for more details.  Each model has between 42,754, and 49,154 parameters. This is a tunable parameter, but was only affected by the number of input signals in this case. 
        
        The prediction performance was evaluated across the test set in terms of Mean Absolute Error (MAE) in the original signal range. This demonstrates that the models performed well in terms of pure predictive error:
        \begin{itemize}
            \item The N1 model achieved an MAE of 0.085 (\% of max speed) 
            \item The P30 model achieved an MAE of 0.715psi
            \item The TGT model achieved an MAE of 1.754°K
            \item The T30 model achieved an MAE of 0.872°K
        \end{itemize}

        \begin{figure}[hbtp] 
            \includegraphics[width=0.49\textwidth]{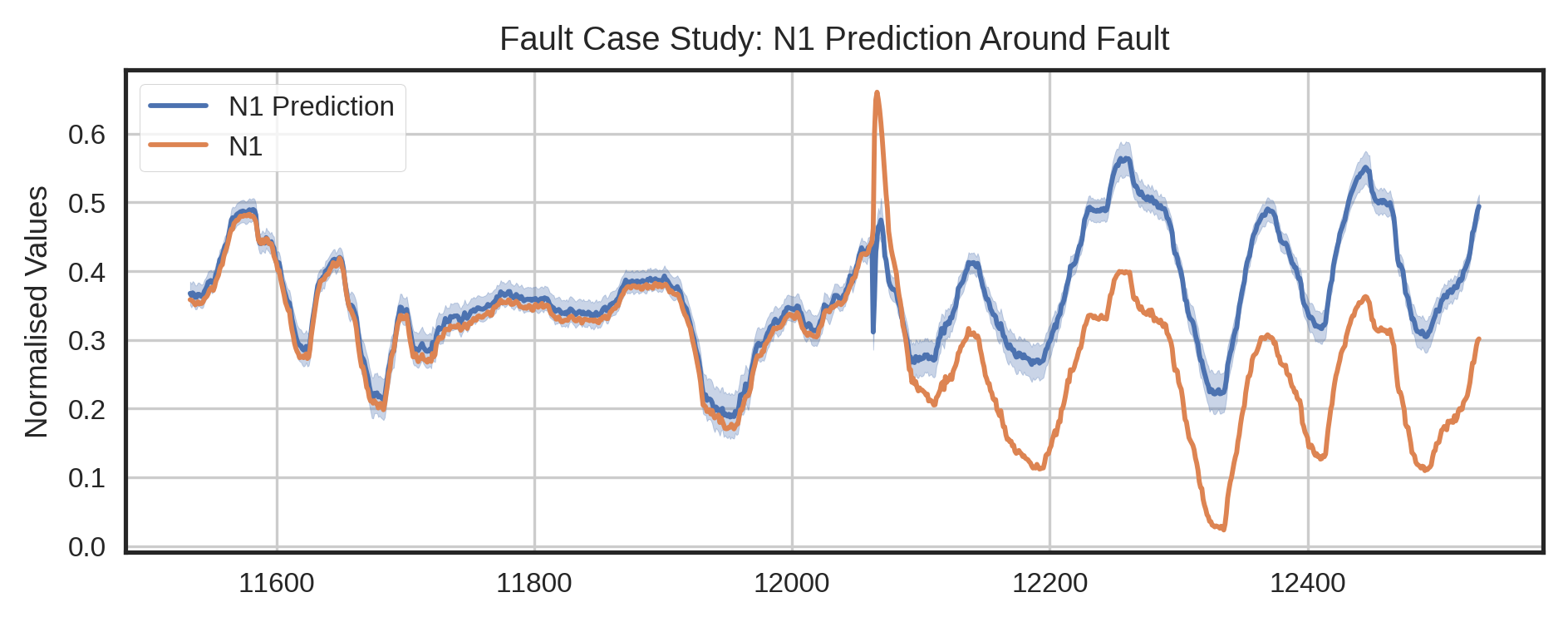}
            \includegraphics[width=0.49\textwidth]{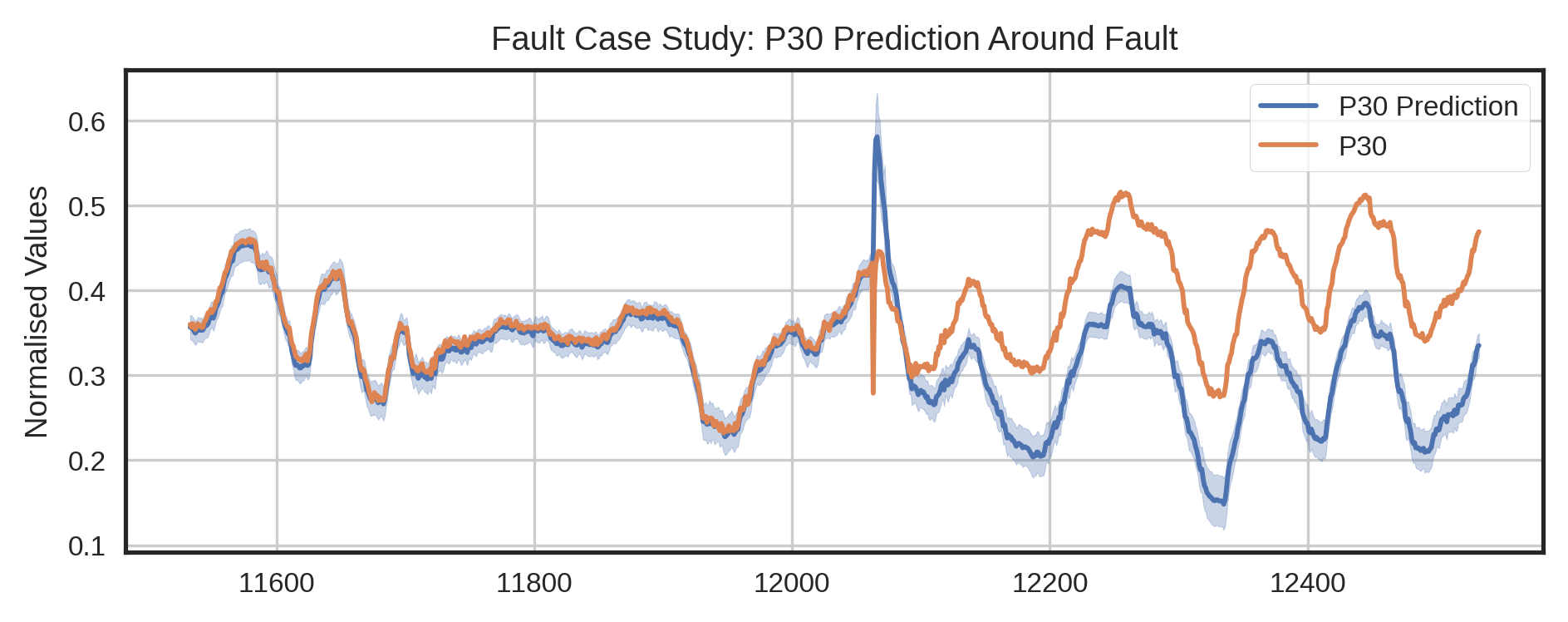}
            \includegraphics[width=0.49\textwidth]{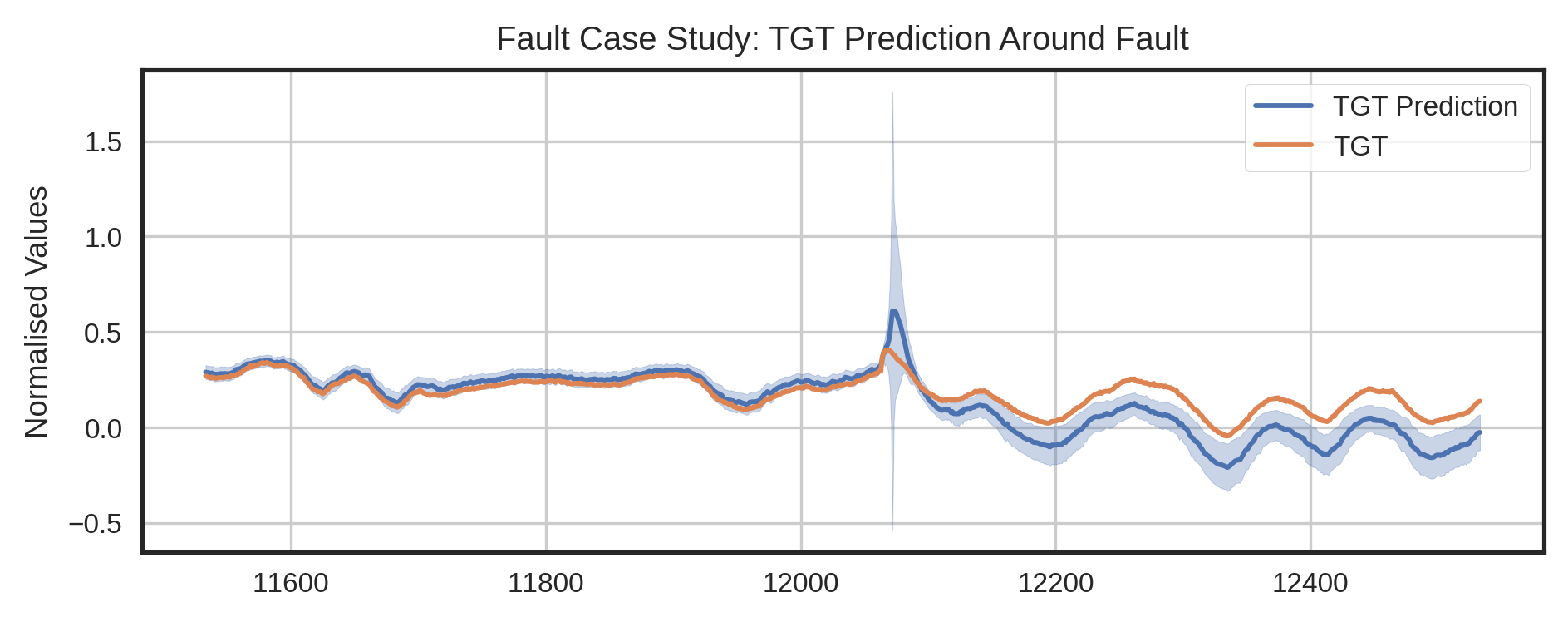}
            \includegraphics[width=0.49\textwidth]{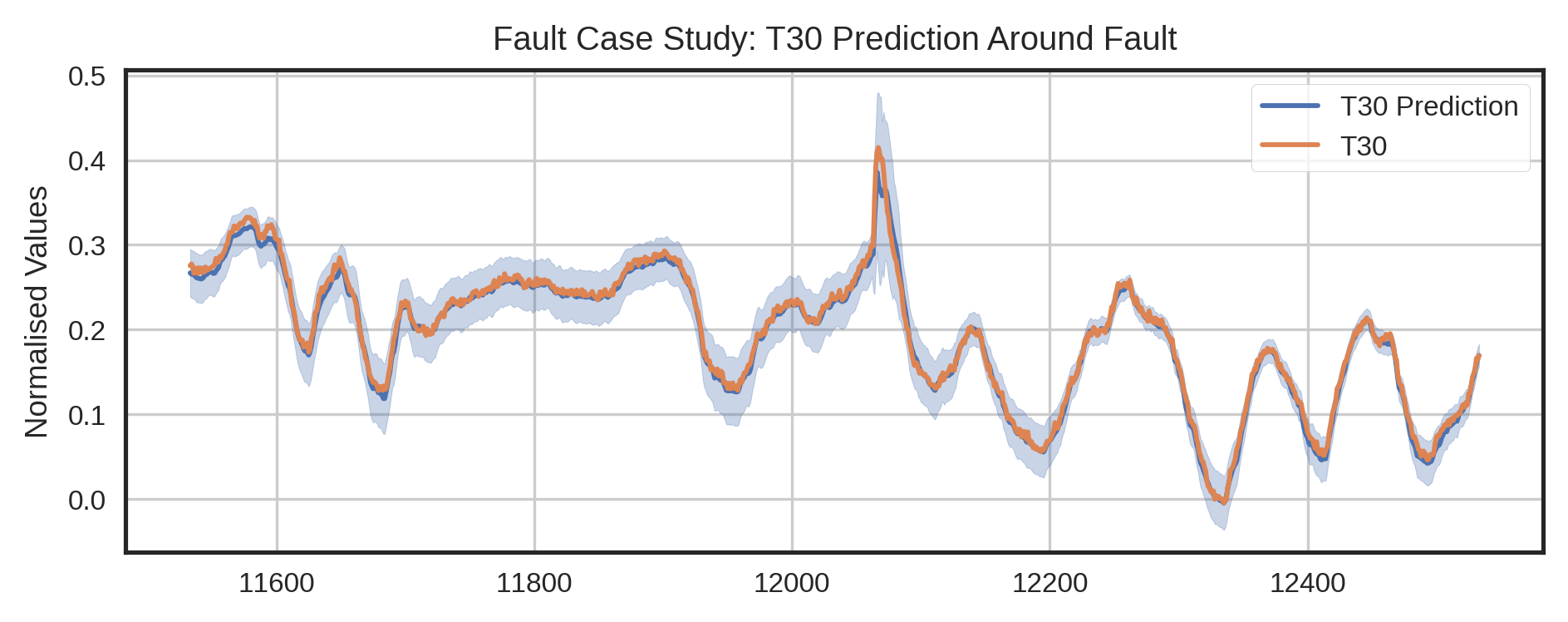}
            \includegraphics[width=0.49\textwidth]{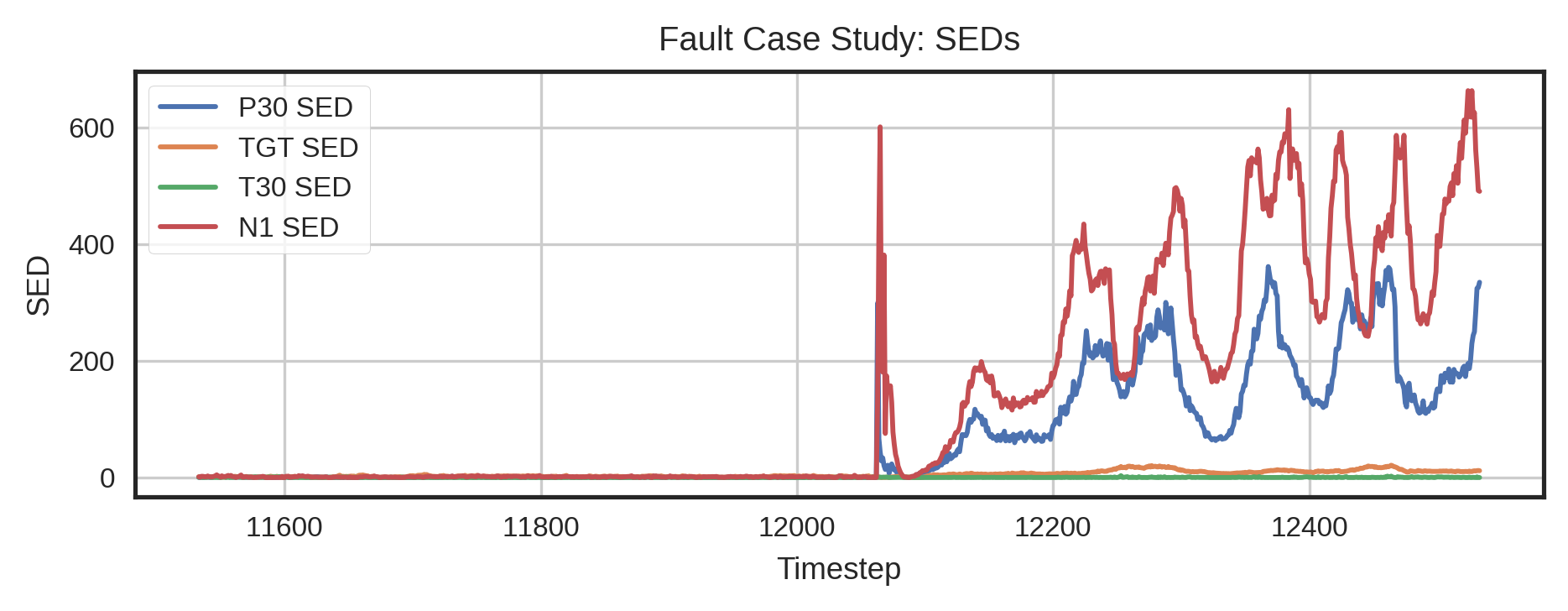}
            \caption{Top 4 graphs: Prediction around the fault, in descending order of average Standardised Euclidean Distance (SED). Highlighted regions indicates a 95\% confidence interval (predicted).
            Bottom graph: SED of each model around the fault.}
            \label{fig:p30_pred}
        \end{figure}

        Several interesting properties of the system are observed from the models' signal predictions at the time region around the known fault, see Figure \ref{fig:p30_pred}. Firstly, the  fault, and the ensuing anomalous behaviour, is clearly detected by the system.  Secondly, the dominant SED (highest magnitude) for this fault is observed in the model predicting N1, despite the fault being most prominent in the P30  signal. This indicates that it is possible for the system to detect anomalous behaviour in signals not being directly predicted. Further, this indicates that the system could monitor a wide range of signals with a small number of models if the set of inputs and outputs is well designed.
        
        In contrast, the T30 model (which also uses P30 as an input) is significantly more accurate in prediction, and therefore rates the data as less anomalous.  This indicates that during the fault, T30 still maintains the same relationship with P30, unlike the other signals. It also indicates that, while using a signal as a model input may allow detection of faults that manifest in that signal, it does not guarantee it.
        
        
        Thirdly, the confidence intervals for all the models remain self-similar, with the exception of the P30 and TGT model during the initial spike. This indicates that the training data covers a sufficient range of operating conditions such that data in this flight is expected to be accurately predicted. The large confidence interval around the initial spike in these models demonstrates that this is the only section (from those tested) that includes data outside  previously seen ranges. A large confidence interval may be an indication that something unusual is occurring, however, since these regions do not appear in  training data, it is not reasonable for the system to identify them as highly anomalous unless the error is also very high.

\begin{figure}[hbtp] 
            \includegraphics[width=0.49\textwidth]{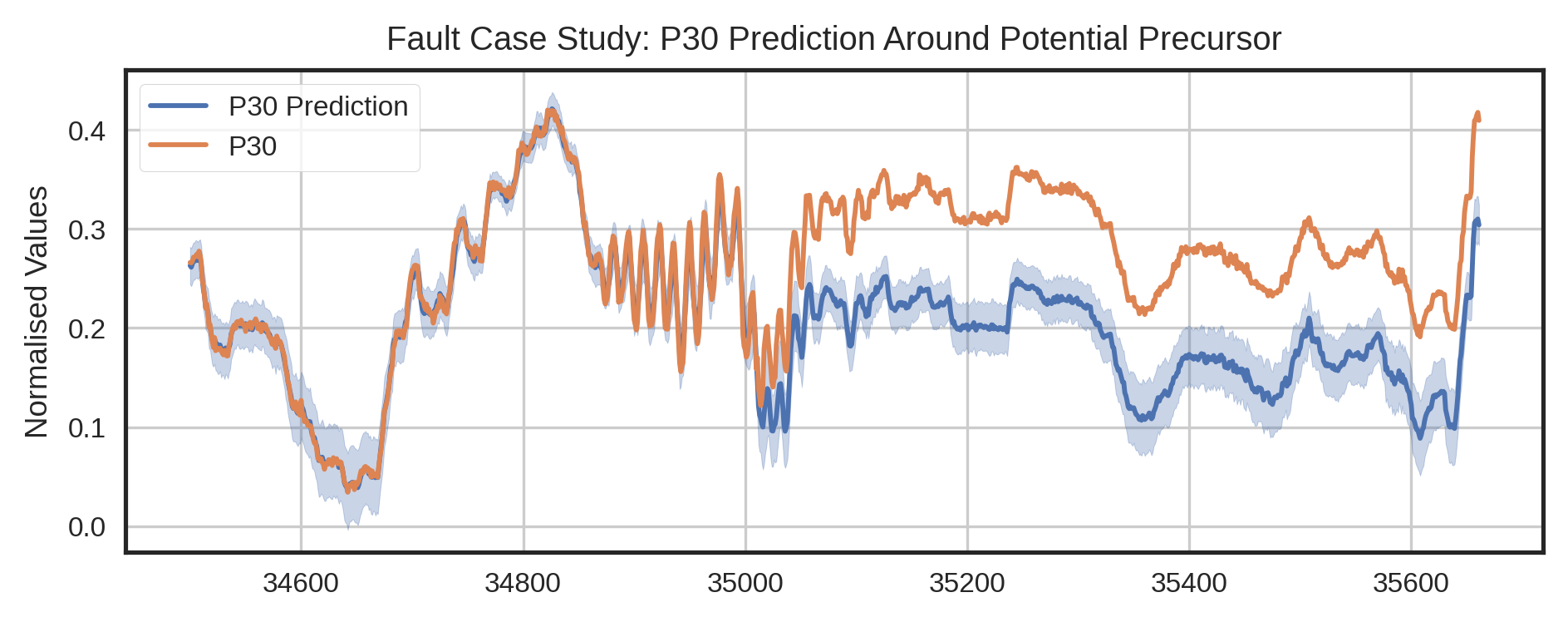}
            \includegraphics[width=0.49\textwidth]{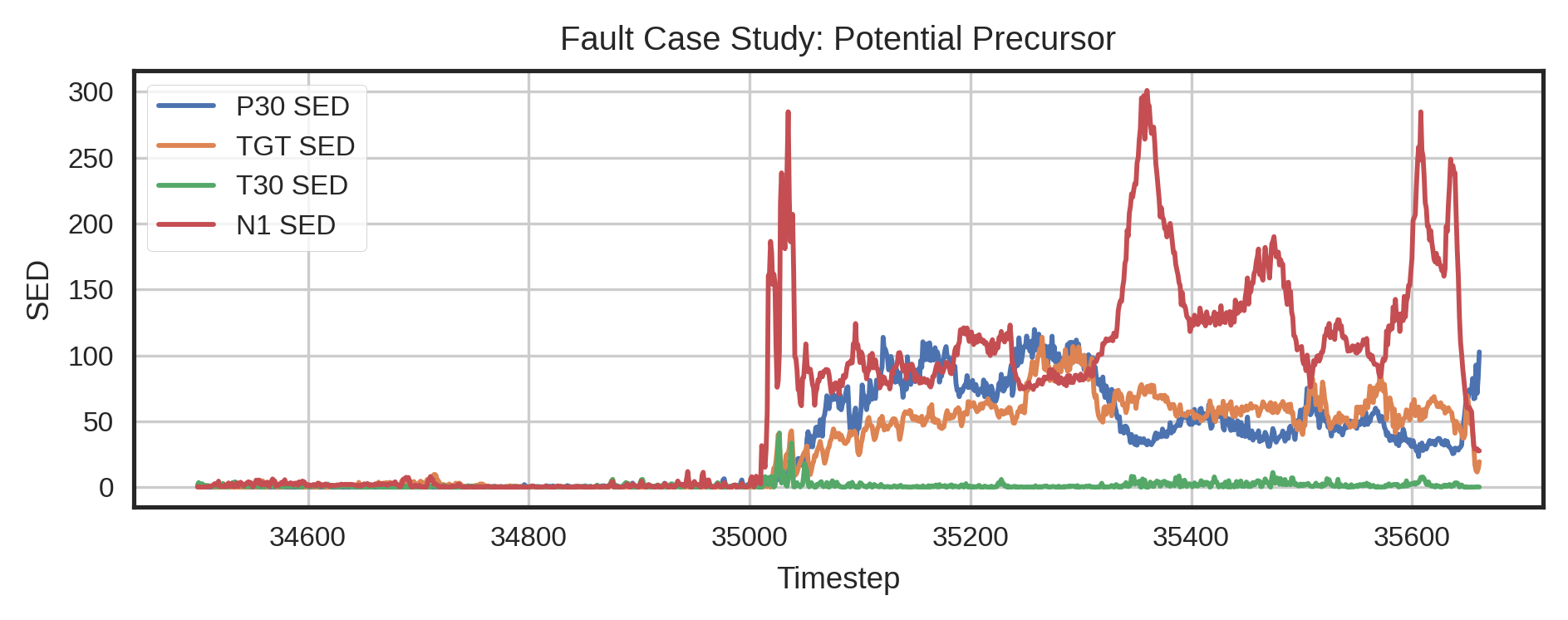}
            \caption{Potential precursor fault, 3 flights before known fault.}
            \label{fig:p30_precursor}
        \end{figure}

        The fault shown in Figures \ref{fig:p30_anomaly} and \ref{fig:p30_pred} is the only fault in the data that has been previously detected, however , the system was able to identify multiple potential precursor events that show the same signature as the known fault, but without the initial spike. These were detected in the flight prior, 3 flights prior, and 6 flights prior, to the known fault. These events were not detected by current health monitoring systems, however, comparison to the other (assumed correctly functioning) engine on the aircraft, they present the same fault signature \cite{jacobsInterengineVariationAnalysis2018}. This indicates that they are highly likely to be related. The system's ability to detect them lends further credence to its effectiveness. An example from the flight 3 flights before the known fault is shown in Figure \ref{fig:p30_precursor}.
        
\begin{figure}[hbtp] 
            \includegraphics[width=0.49\textwidth]{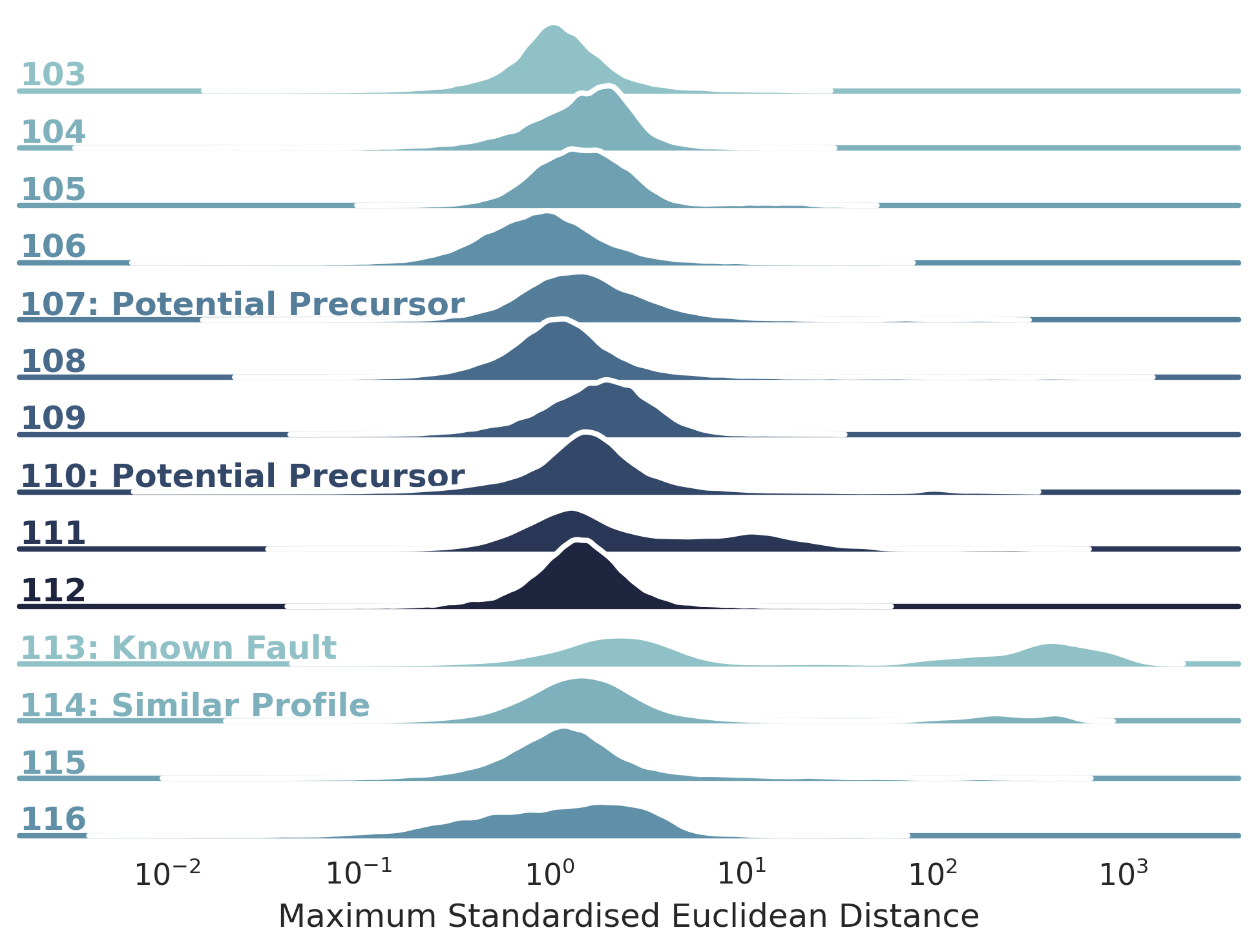}
            \caption{MSED Distribution for test flights. Known fault has a large amount of mass at a much higher MSED than other flights. Potential precursors show smaller mass at similar MSED ($>100$). Flight 111 also exhibits an odd profile due to irregularity in the shaft speed signal not found in other flights.}
            \label{fig:msed_dist_p30}
        \end{figure}

        The MSED for the flight containing the known fault (flight 113), as well as the following flights, are observed to be distributed differently from the normal flights, see Figure \ref{fig:msed_dist_p30}. The precursor faults also show a similar shaped distribution with a second peak at MSED $> 100$. This indicates that the system is correctly prioritising unusual data on real flight data.
        
        Flight 111 also demonstrates an unusual distribution; this is due to unusual behaviour in the shaft speed signal which is not found in other flights. It is unclear whether this is another fault or an unusual but harmless behaviour. This highlights the benefit of data prioritisation, over binary labelling since an operator can review the data to make the final determination on this sort of edge case.
        
        
    \subsection{Real-time Performance} \label{sec:realtime}
        The system was deployed on a Xilinx Zynq XC7Z020-2CLG484I, an industrial grade device based on an ARM Cortex A7 microprocessor, and on-wing system target. In order to test real-time operation, the models designed above were deployed to the device and constrained to $<=20\%$ of CPU time. A synthetic data set of 25 signals over 100,000 time points was sent to the device and an output buffer of size 50 windows was used so that the 50 most unusual windows would be returned per ground data transmission. 
        
        
        The actual processing time for each run was recorded; a run consisted of all currently uploaded models being run once, the MSED being determined and the output buffer being updated if necessary. The other primary resource-intensive operation, input retrieval and clean-up, was ignored since its computational expense is heavily dependent on the data source. Although, in our application its expense was orders of magnitude smaller than that of a run ($\sim25\mu s$).
        
\begin{figure}[ht] 
            \includegraphics[width=0.49\textwidth]{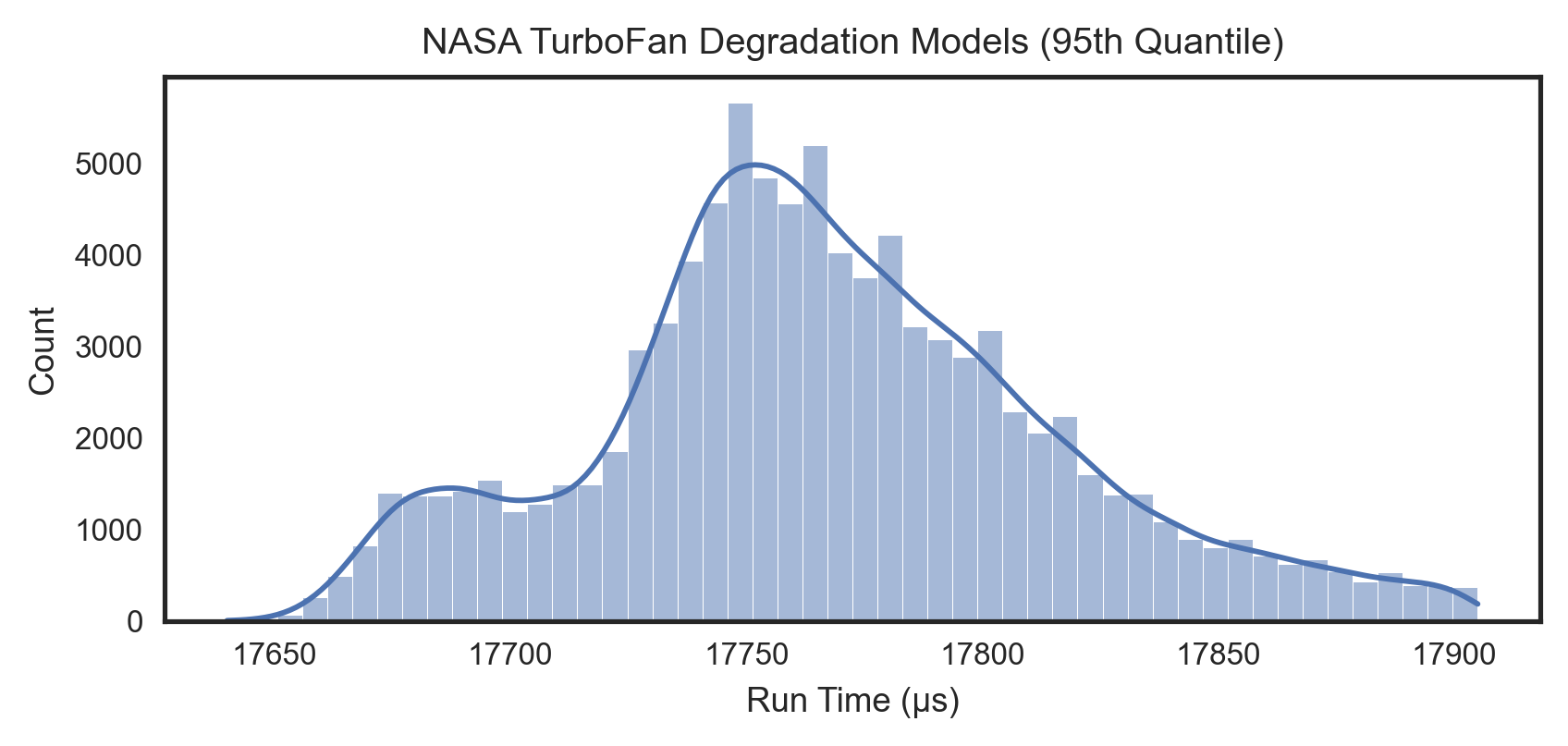}
            \caption{95th quantile of run durations for the NASA TurboFan Degradation data. Smaller peak likely caused by cache hits.}
            \label{fig:nasa_runtime}
        \end{figure}

\begin{figure}[ht] 
            \includegraphics[width=0.49\textwidth]{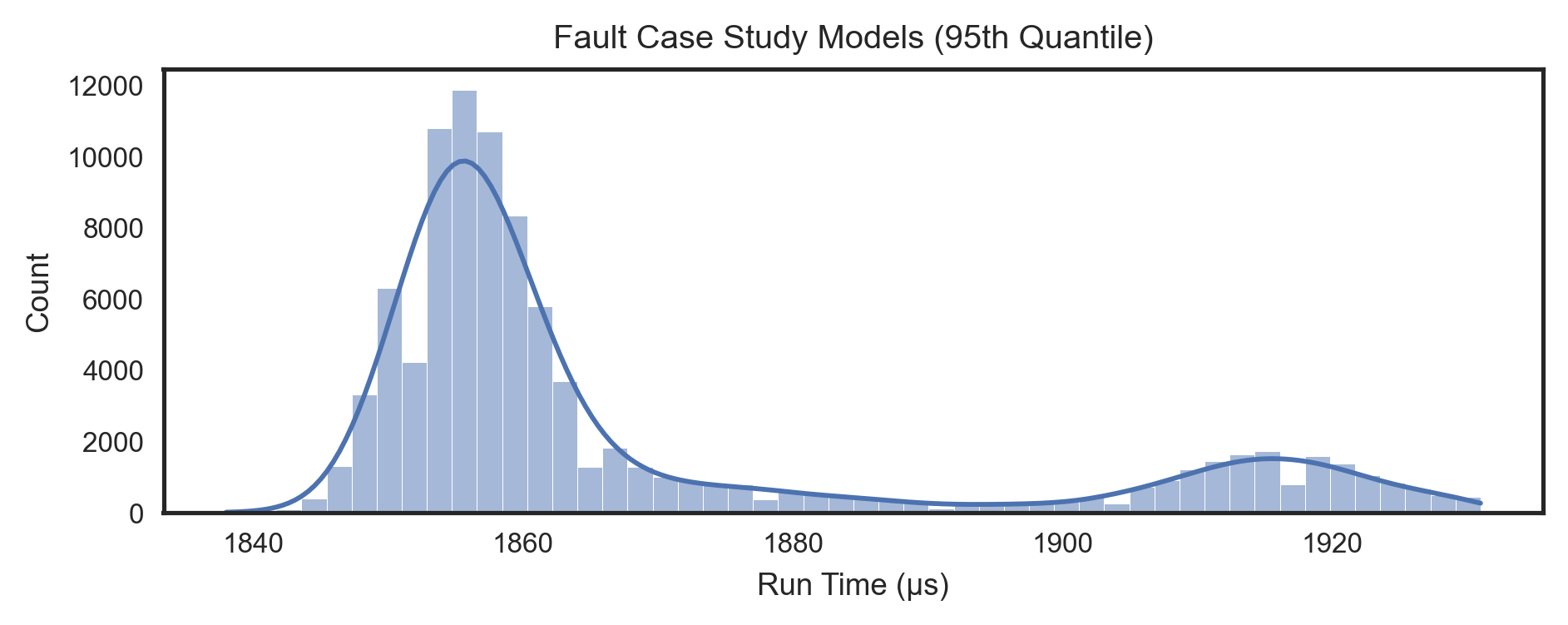}
            \caption{95th quantile of run durations for the Fault Case Study data. Smaller number of models and overall smaller model size relative to NASA data lead to the lower average run duration and likely the reversal of shape due to more cache hits.}
            \label{fig:p30_runtime}
        \end{figure}

        Figure \ref{fig:nasa_runtime} shows the distribution of run durations when the 20 models used to evaluate the NASA data set were used and Figure \ref{fig:p30_runtime} when the 4 models used to evaluate the In-Service Fault Case Study data set were used.
        
        Both histograms have a long tail caused by the fact that the software runs on a non-real-time operating system (unpredictable system interrupts) which is why they are curtailed at the 95th quantile. These results show that the system is capable of operating in real-time e.g. all 20 of the NASA models could be run simultaneously on data at up to $\sim50Hz$ and the Fault Case Study models could be run at up to $\sim500Hz$, including all overhead. 
        
\section{Conclusion}
    We have presented a data-driven anomaly detection system that runs in real-time on low-power embedded hardware. The system prioritises data based on the Maximum Standardised Euclidean Distance of a flexibly sized ensemble of models. A flexible neural network design is presented that is used to build models for the ensemble from any input-output set and window length without the need for architecture changes. The efficacy of the system has been demonstrated on both real time-series and synthetic snapshot data, plus its real-time operation on embedded hardware has been proven and characterised.

    The system represents a practically realisable, computationally inexpensive method for detection of unusual behaviour in real-time. Further the architecture is flexible and allows for deployment on new systems with minimal need for in-depth domain knowledge.
    
    The system will be deployed on-board in an upcoming flight test program. Future work will focus on analysis of the data returned by this test and how the process described here-in can be applied to higher frequency data.



\section{Bibliography}
\bibliographystyle{unsrt}
\bibliography{main}

\appendix
\section{Appendix 1: Network Reproduction} \label{sec:app1}

The loss function (Equation \ref{eq:loss}) was simplified and reordered for batch-wise implementation since the model's confidence output $\alpha = log(\sigma)$ rather than $\sigma$: 
        
\begin{equation}
    \mathcal{LL} = \frac{1}{2} \sum\limits_{n=1}^{N} \Big(\frac{y_{n} - \mu}{e^{\alpha_n}} \Big)^{2} - \sum\limits_{n=1}^{N} \alpha_{n} - \frac{1}{2} log(2\pi)
\end{equation}
where $N$ is the batch size. The final term is constant and could therefore be dropped for computational efficiency, although it was kept during our work to keep the meaning of the equation intact.

Important network hyper-parameters not mentioned elsewhere:
\begin{itemize}
    \item LReLU $\alpha = 0.3$
    \item Dropout Rate = $0.5$
    \item Convolutional layers
    \begin{itemize}
        \item Each used 64 filters in the NASA study
        \item Each used 32 filters in the Fault Case Study
        \begin{itemize}
            \item An optimisation found to cause minimal performance loss while reducing computation
        \end{itemize}
        \item Padding irrelevant for 1x1 blocks, \enquote{same} used internally
        \item No padding for Temporal and Spatial Fold blocks
        \begin{itemize}
            \item Used to reduce the size of the input
            \item Equivalent to \enquote{valid} padding
        \end{itemize}
    \end{itemize}
    \item Each dense layer had 64 hidden units for the signals tested
    \item Early stopping patience was 10 epochs without improvement
    \item RADAM parameters (adjusted based on validation data for each signal)
    \begin{itemize}
        \item Learning rate $1e-3$ to $1e-5$
        \item Min learning rate $1e-5$ to $1e-7$
        \item Total steps $50,000$ to $200,000$
        \item Warm-up proportion $0.1$ to $0.5$
    \end{itemize}
\end{itemize}

\begin{figure}[hbtp] 
    \centering
    \includegraphics[width=0.4\textwidth]{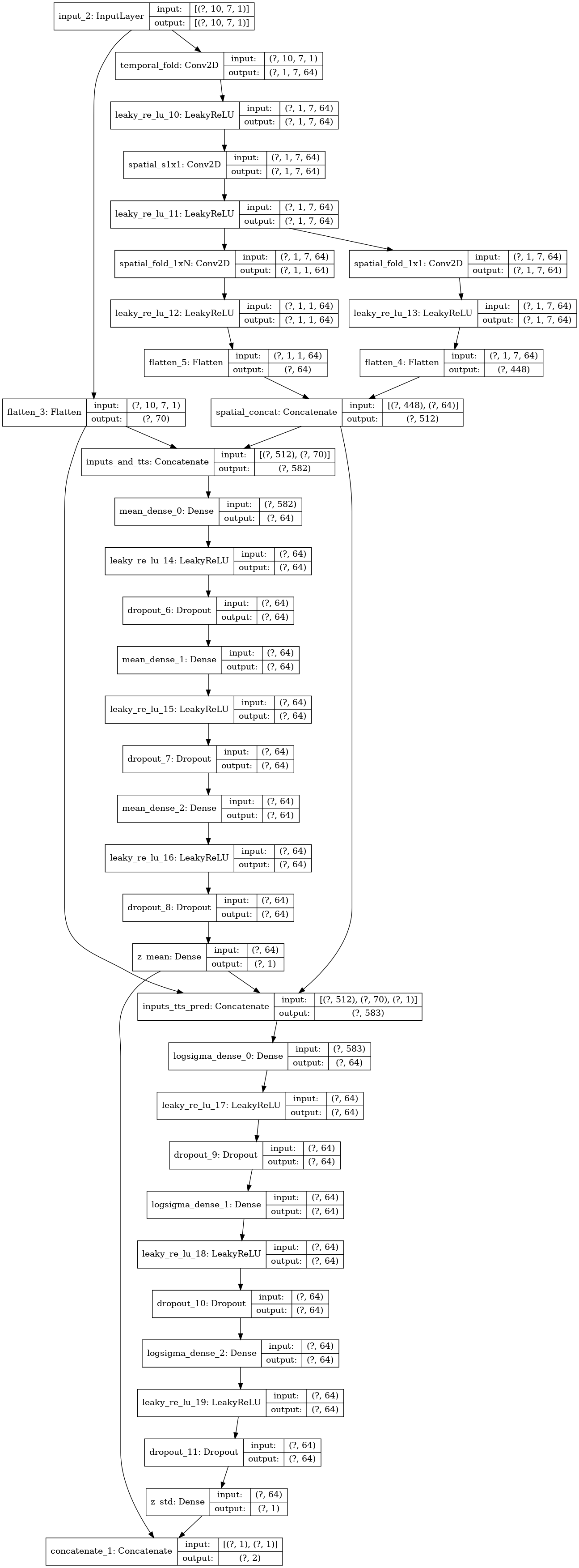}
    \caption{Neural network architecture with 7 input signals and a 10 second window.}
    \label{fig:nn_arch}
\end{figure}

An example implementation of the architecture used on our testing for 7 inputs, looking at 10 seconds of data per window is shown in Figure \ref{fig:nn_arch}. 

\section{Appendix 2: Notes on Data Preprocessing} \label{sec:app2}
The same process was applied to each dataset:
\begin{itemize}
    \item Data was scaled to the range [0, 1]
    \begin{itemize}
        \item Based on 2\% to 98\% quantiles in the training set
    \end{itemize}
\end{itemize}

\section{Appendix 3: In-Service Fault Case Study} \label{sec:app3}
The four models were:
\begin{itemize}
    \item P30 (internal pressure) from outside temperature, outside pressure, and N1
    \item TGT (Turbine Gas Temperature) from outside temperature, outside pressure, P30, and N1
    \item T30 (internal temperature) from outside temperature, outside pressure, P30, and N1
    \item N1 (speed of low speed shaft) from outside temperature, outside pressure, and P30
\end{itemize}
\end{document}